\documentclass[11pt]{article}
\usepackage{geometry}
\usepackage{coling2020}
\usepackage{times}
\usepackage{url}
\usepackage{latexsym}
\usepackage{microtype}
\usepackage{enumitem}
\usepackage{graphicx}
\usepackage{hyperref}
\usepackage{color}
\usepackage{soul}
\usepackage{enumitem}
\usepackage{tikz-dependency}

\usepackage{cleveref}
\crefname{section}{§}{§§}
\Crefname{section}{§}{§§}

\colingfinalcopy 

\usepackage{soul}

\title{IITK-RSA at SemEval-2020 Task 5: Detecting Counterfactuals}

\author{Anirudh Anil Ojha$^{*}$ \qquad    
  Rohin Garg$^{*}$  \qquad  
  Shashank Gupta\thanks{\quad Authors equally contributed  to this work.} \qquad
  Ashutosh Modi  \\
{Indian Institute of Technology Kanpur (IITK)} \\
  {\tt \{aaojha, sronin, gshasha\}@iitk.ac.in}  \\
  {\tt ashutoshm@cse.iitk.ac.in}  \\
}
\date{}
\begin{document}
\maketitle
\begin{abstract}
This paper describes our efforts in tackling Task 5 \cite{yang-2020-semeval-task5} of SemEval-2020.
The task involved detecting a class of textual expressions known as counterfactuals and separating them into their constituent elements. Counterfactual statements describe events that have not or could not have occurred and the possible implications of such events. While counterfactual reasoning is natural for humans, understanding these expressions is difficult for artificial agents due to a variety of linguistic subtleties. Our final submitted approaches were an ensemble of various fine-tuned transformer-based and CNN-based models for the first subtask and a transformer model with dependency tree information for the second subtask. We ranked 4$^{th}$ and 9$^{th}$ in the overall leaderboard. We also explored various other approaches that involved the use of classical methods, other neural architectures and the incorporation of different linguistic features.
\end{abstract}



\section{Introduction}
\label{intro}

%
%
\blfootnote{
    %
    %
    %
    
    \hspace{-0.65cm}  
    This work is licensed under a Creative Commons 
    Attribution 4.0 International Licence.
    Licence details:
    \url{http://creativecommons.org/licenses/by/4.0/}.
     
    %
}

Counterfactuals \cite{sep-counterfactuals} are statements that describe events that did not (or cannot) happen, which implies that their truth cannot be empirically verified. They may be divided into two major categories \cite{iatridou2000grammatical}. In the first category, the statement discusses possible outcomes had the event in question actually happened, and thus is called a \emph{counterfactual conditional}. Specifically, it contains a logical segment equivalent to ``if A, then B" where A did not occur. The ``if A" part is called the \emph{antecedent} and the ``then B" is called the \emph{consequent}. An example of this is ``If we had changed the battery on time, the car wouldn't have broken down." 
A statement of the second type is called a \emph{counterfactual wish}. They also describe alternate realities, but do not talk about outcomes and do not display a conditional structure. They are called so because they often display `wish' vocabulary e.g. ``I wish there was an easy way to categorise counterfactuals!" Here, the segment referring to the fictional event is called the antecedent, and there is no consequent. While there are other ways to categorise these logical forms, we believe that this system covers most of the examples seen in the dataset and provides clear intuition about their general structure. Readers interested in a more rigorous treatment may refer to \cite{lewis2013counterfactuals}. 

The ability to reason via counterfactuals is considered a higher form of intellect, as one must extrapolate the truth using causal relationships, a commonsense understanding of the world and other related events. Hence, for artificially intelligent systems, counterfactual reasoning is critical for generalisation and explainability. In terms of more concrete applications, it has been hypothesised to be important for natural language understanding, goal determination (a major challenge in modern reinforcement learning) and  error analysis \cite{ginsberg1986counterfactuals}. Recently, \newcite{mothilal2020explaining} have worked on systems that generate counterfactual explanations for why classifiers behave the way they do on given examples. 

We tried a variety of approaches on this task, including classical machine learning, CNN-based models, BiLSTM-based models and transformer-based models. We found that the transformer-based neural models (introduced by \newcite{vaswani2017attention}) consistently outperformed their competitors. We obtained an F1 score of 89.3\% on the test data of the first subtask and 48.3\% on the second subtask, ranking our system 4th and 9th respectively. Our implementation is available via  Github.\footnote{\href{https://github.com/gargrohin/counterfactuals-nlp}{https://github.com/gargrohin/counterfactuals-nlp}}

\section{Brief Task and Dataset Description}

The overall task consists of two subtasks:
\begin{itemize}[noitemsep,topsep=4pt]
    \item Detecting counterfactual statements
    \item Parsing such statements into antecedent and consequent (if present)
\end{itemize}

The first may be modelled as a binary classification problem and the second as a sequence-to-sequence labelling problem. The training set was provided by the task organisers. The examples were segments from news articles in English and were quite varied in length, from around 5 words to around 400 words. There were 13000 examples for the first subtask, with an $\sim$88:12 negative-positive split, and 3500 for the second subtask. As the datasets were imbalanced, our evaluation metrics were precision, recall, F1-score and exact match (exclusively for the second subtask). The test set had 7000 sentences for the first subtask and 1950 for the second. Some samples from the training set are given in Tables \ref{table:ex_1} and \ref{table:ex_2}. For more details, please refer to \newcite{yang-2020-semeval-task5}. 

\begin{table}
\begin{center}
\begin{tabular}{|p{13cm}|p{1.9cm}|}
\hline \bf Sentence & \bf Gold Label \\ \hline
I wish it had more. & True \\
Even if the next government awards the NHS an annual funding increase of 1.5 per cent over inflation, which is no simpler than giving everyone good enough medical care whenever they need it, the service would have still had to make annual efficiency savings of 2-3 per cent a year. & True \\ 
If Mr McCain is in charge, his record of bipartisan outreach will stand him in good stead; Mr Obama will be able to rely on solid majorities in Congress. & False \\
If your device is locked and no one can discern the security code, you may not need a patent on it. & False \\
\hline
\end{tabular}
\end{center}
\caption{\label{table:ex_1} Samples from the training set of the first subtask. Gold Label True means statement is a counterfacual.}
\end{table}

\begin{table}
\begin{center}
\begin{tabular}{|p{5.5cm}|p{4.5cm}|p{4.5cm}|}
\hline \bf Sentence & \bf Antecedent & \bf Consequent\\ \hline
The GOP's malignant amnesia regarding the economy would be hilarious were it not for the wreckage they caused. & were it not for the wreckage they caused & The GOP's malignant amnesia regarding the economy would be hilarious \\
"I wish there was no limit on the number of groups you could join because there are so many good ones," Larsen said. & I wish there was no limit on the number of groups you could join & \\ 
Duda said that if Netanyahu had said what was originally reported, "Israel would not be a good place to meet in spite of the previous arrangements". & Duda said that if Netanyahu had said what was originally reported & Israel would not be a good place to meet in spite of the previous arrangements \\
\hline
\end{tabular}
\end{center}
\caption{\label{table:ex_2} Samples from the training set of the second subtask. The objective is to mark the characters in the antecedent and the consequent as separate labels.}
\end{table}

\section{Previous and Related Work}
\label{srel}

Research on counterfactuals is a relatively new area in the NLP community, and consequently, limited research literature exists. The closest work is by \newcite{son2017recognizing} which aimed to detect counterfactual elements in tweets. They identify a set of rules to filter such examples based on grammatical form (sentences containing ``wish", ``should have", etc.) and train an SVM to classify filtered sentences on the basis of POS tags and n-gram features. While their method works well on tweets, we found that these rules had limited coverage and were not effective in filtering long sentences like those in our dataset. The filtering also removes several examples, which makes it harder to train further classifiers in an already data-constrained setting.

\newcite{iatridou2000grammatical} provides a thorough treatment of counterfactuals in terms of grammar. However, the features and verb forms they identify and use are quite complex, like fake tenses. Such morphology is currently undetectable by even state-of-the-art libraries like spaCy\footnote{\href{https://spacy.io}{https://spacy.io}}, which makes this system hard to implement with computational models without appropriate data. 
Nevertheless, the paper does allow one to construct certain useful grammatical rules for inference. 

Counterfactual implication and causality are closely linked \cite{morgan2015counterfactuals}. Counterfactual conditionals often talk about direct results of the presence (or absence) of a particular factor e.g. ``If the gas hadn't run out, we would have been able to cook dinner." Even when they do not directly encode such relationships - one may consider sentences like ``Even if you were the last person on the planet, I wouldn't go out with you!" - they often imply the existence of other factors that drive the relationship. Thus, from a linguistic perspective, the vocabulary and structure seen in causal sentences and that seen in counterfactual sentences have a high degree of overlap. 
Hence, we believe systems to detect and parse causal relationships would perform well on this task as well. With respect to the first subtask, fine-tuning neural models using information-rich word embeddings seems to form the state-of-the-art \cite{kyriakakis2019transfer} and forms the backbone of our submitted approach as well. The second subtask bears a resemblance to relation-entity extraction (of which cause-effect may be considered a specific case), which has also been recently dominated by neural models \cite{li2020downstream,soares2019matching}. We found span-based \cite{joshi2020spanbert} and discourse parsing-based \cite{son2018causal} approaches particularly interesting with regards to our task.

\section{Our Approaches}

\subsection{Subtask 1}

\subsubsection{Classical machine learning}
\label{sssec:cml}

We tried SVMs and gradient-boosted random forests on linguistic feature-based (POS tags, verb tense and aspect information, etc.) representations of the samples. The baseline provided by the organisers was an SVM classifier on TF-IDF features with stop words removed. We felt that this would result in the removal and down-weighting of terms critical for identifying counterfactuality, like `should have' and `would have'. Indeed, we found that performance improved remarkably by discarding TF-IDF features and allowing stop words. This implies that the way that closed class words are used in a sample is particularly informative in this task. Even using simple unigram word features provided decent results, which suggests that there are specific words whose presence in itself is useful for detecting counterfactuality. Thus, we used these features in our neural model as well (see \cref{sssec:trans_s1} c)). 
Even though the results improved by the progressive addition of linguistic features, the F1 score eventually stagnated at around 65\%.

We then added a regex-based filtering step, where we divided the samples based on certain typical counterfactual forms (inspired by the work of \newcite{iatridou2000grammatical}) and classified them separately. We used four primary forms - sentences containing an `if' then a modal verb, sentences containing a modal verb then an `if', sentences containing the word `wish', and the leftovers. In the first two, we wished to capture verb forms like `would have' and `could have', which we found to have a strong correlation with counterfactuality. We found the modal-if form the hardest to classify, with an F1 of around 52\%. The most notable improvements were on the `wish' form where the F1 was around 90\%, which hints that counterfactual wishes are the easiest to identify. Classifiers for the other two categories obtained an F1 of around 62-65\%, similar to what we obtained before filtering. 

\subsubsection{CNN with GloVE and Word2Vec}
\label{sssec:cnn}

Inspired by \cite{kim-2014-convolutional}, we replaced the words with pretrained word embeddings and applied a CNN-based classification head. For each sentence, the words were tokenised and replaced with their embedding. Thus, a matrix was formed with dimension (embedding length $\times$ maximum sentence length). We experimented with multiple kernel sizes and with static and non-static versions of word-embeddings, as in the paper. We finally settled on 100 kernels of sizes 3, 4 and 5 each. There was a max-pooling layer before the fully-connected layer. We abandoned the non-static embeddings layer because training the embeddings resulted in overfitting. GloVe \cite{pennington2014glove} embeddings performed 3-4\% better F1-wise than Word2Vec \cite{mikolov2013distributed} embeddings as seen in Table  \ref{table:res_1}.

\subsubsection{Transformer-based models}
\label{sssec:trans_s1}
Introduced by \newcite{vaswani2017attention}, transformer-based models have recently reported state-of-the-art (SOTA) results on a variety of NLP tasks, and we found the same to be true for our task as well. Since our dataset was small, we experimented with transfer learning on three state-of-the-art variants - BERT \cite{devlin2018bert}, RoBERTa \cite{liu2019roberta} and XLNet \cite{yang2019xlnet}. We tried the following modifications (exact values reported in Table \ref{table:res_1} and \cref{sec:result}):

\begin{enumerate}[noitemsep,topsep=4pt,label=\alph*)]
    \item \textbf{Fully-connected head:} This involved combining the word-level embeddings generated by the transformer into a sentence-level embedding and then applying a fully-connected layer for classification. We experimented with various ways of producing the sentence embedding, such as taking the mean, max-pooling (taking the maximum value over each dimension) and selecting particular embeddings. We found that taking last word embedding (the [CLS] token in BERT and RoBERTa) gave us the best results. We applied a tanh activation between the sentence embedding and the linear layer. `Large' variants performed around 5\% better on F1 than their `base' counterparts. Despite the simplicity of this approach, it was second only to the ensembles (see section \cref{sssec:ens}). We also tried to fine-tune these models on subsets of the data after applying a filtering step as in \cref{sssec:cml}. However, the F1 score stagnated around 80\%, probably because of insufficient data.
    \item \textbf{CNN head:} Instead of obtaining a sentence embedding, we directly applied a CNN head on the word-embedding matrix from the transformer. This is similar to what we did in the case of vanilla CNN model (\cref{sssec:cnn}), and we decided to apply the same final CNN head that we used in that section. While this did provide about 3-4\% better F1 than the `base' versions with the fully-connected heads, it was still prone to overfitting and could not outperform the `large' versions.
    \item \textbf{With linguistic features:} Following up on our observations in \cref{sssec:cml}, we appended linguistic features like word n-grams and POS n-grams to the transformer embeddings before passing it through a fully connected layer. We experimented with unigrams, bigrams and trigrams of words and POS tags, and eventually settled on the top 1000 trigram features of each. This addition stabilised training and improved performance on `base' models by 3-4\% F1, but could not outperform our fine-tuned `large' models. 
\end{enumerate}

\subsubsection{Discourse parsing-based approaches}

We took inspiration from \newcite{son2018causal} and parsed the samples into discourse arguments with PDTB-style methods. 
We were motivated by two key observations - there was a high degree of overlap between relevant discourse arguments and sections encoding counterfactuality (antecedent and consequent), and in longer examples, only a small part of the sentence was counterfactual. Thus, we hoped that this parsing could help the system better filter out the antecedent and consequent in longer examples and focus on understanding their specific structure.

We first used spaCy's dependency parser on the samples, and then used this information along with the presence of known discourse connectives like `even though' and `because' to obtain discourse arguments. After parsing, we tried two architectures. In the first, we obtained word-level GloVe embeddings as in \cref{sssec:cnn} and applied a hierarchical BiLSTM network. The first BiLSTM network was applied at the word level, over each discourse argument, thus giving us a discourse-level embedding in the form of the final hidden state. These discourse-level embeddings were taken sequentially and fed into another BiLSTM. The final hidden state thus generated was fed into a fully-connected layer. 

In the second approach, we used a RoBERTa model to obtain the discourse-level embedding, similar to how we obtained the sentence-level embedding in \cref{sssec:trans_s1} a). Each discourse argument was limited to 32 words, with a maximum of 8 discourse arguments. These discourse embeddings were recurrently fed into a BiLSTM as before. We initialised the discourse-level BiLSTM with the sentence-level embedding of the samples obtained as in \cref{sssec:trans_s1} a). Hence, the hidden state dimension of the BiLSTM was the same as that of the transformer embedding. While we were able to train this with `base' transformers, the `large' variants took too long. 

Despite our intuition, GloVe embeddings and the hierarchical BiLSTM performed slightly worse than the Glove $+$ CNN model in \cref{sssec:cnn}. In the case of transformer embeddings with a BiLSTM, we did see an improvement over the results in \cref{sssec:trans_s1} a) on `base' transformers, but not more than what we had already obtained before in \cref{sssec:trans_s1} b) and \cref{sssec:trans_s1} c), suggesting that this also suffer from overfitting due to a large number of parameters. The results are recorded in Table \ref{table:res_1}.

\subsubsection{Ensembles}
\label{sssec:ens}
    We took ensembles of the different models we had trained so far. This was based on our observation that different models failed on different examples, and thus a simple voting algorithm would smooth out inconsistencies. We also observed that RoBERTa models usually had higher recall while XLNet models had higher precision, and thus could complement each other. We could only experiment with hard voting ensembles due to a lack of time - it would be interesting to explore how other methods would perform. This approach yielded the best results overall. We finally submitted two ensembles. The first ensemble model is a combination of (individually fine-tuned as in \cref{sssec:trans_s1} a)) an XLNet large, an XLNet base and a RoBERTa large. The second ensemble model is a combination of a RoBERTa large, an XLNet large, an XLNet base, a BERT base with CNN head, a BERT large and a CNN model with GloVe embeddings. The second model gave the best results on the final test set.
\subsection{Subtask 2}
\subsubsection{Sequence Prediction}
\label{sssec:token}
We tried two approaches while treating this as a sequence prediction task:

\begin{enumerate}[noitemsep,topsep=4pt,label=\alph*)]
    \item \textbf{Token classification using transformers:} Motivated by the performance of transformers in Subtask 1, we experimented with transformer-based models on this subtask as well. We approached the problem like a sequence prediction task similar to named-entity recognition. We tokenised the string into words and gave BIO tags to the words in the antecedent and consequent separately for training. We fine-tuned separate models for detecting antecedent and consequent, as we noticed that the two sections often overlapped (sometimes completely). As RoBERTa gave us the F1-best performance in the first subtask, we experimented with BERT and RoBERTa.

    \item \textbf{End to End Sequence Labelling:} We also modelled the task as an end-to-end sequence labelling problem, where each word was be labelled as an antecedent or a consequent. We used the algorithm described by \cite{ma2016end} in which word representations and character-level representations are used as inputs to a bi-directional LSTM model. The two hidden states for each word were concatenated to form the final output. Finally, a Conditional Random Field (CRF) \cite{lafferty2001conditional} was used for sequence labelling.
\end{enumerate}

\subsubsection{Using dependency tree}
\label{sssec:tree_ant}
An exploratory analysis of the dataset showed that around 60\% of the sentences in the dataset had the word `if'. Out of these, 90\% were in antecedents, thus showing a correlation between the usage of the word `if' and part of the sentence being an antecedent. Given this observation, we tried to use this marker to deal with such antecedents separately.

We separated these sentences and analysed the dependency tree using a dependency visualiser to find a pattern to the antecedents which started with the word `if'. In most cases where it was used, the word `if' was a subordinating conjunction that depended on a noun/verb from an independent clause in the dependency tree. The last word of the antecedent in nearly all the cases was the rightmost leaf node of the sub-tree rooted in that word, i.e. the last word up to which that section of the tree extended. The antecedents extracted from the dependency tree were closest to ground truth, which is why we used them in our final model for the relevant sentences.

\begin{figure}[h]
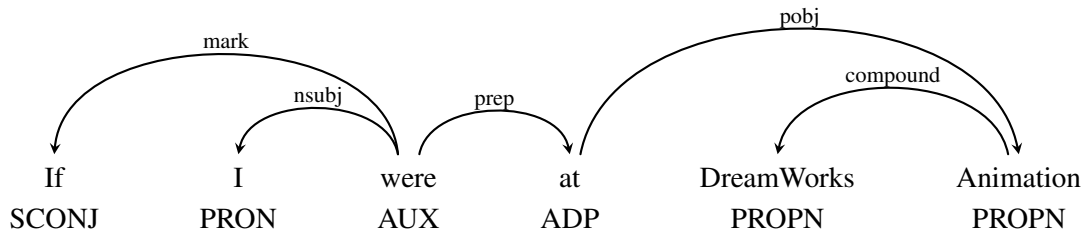

\begin{center}
\begin{dependency}[theme = simple, label style={font=\large}, edge vertical padding=0.2em, edge style=thick]
   \begin{deptext}[column sep=3em, row sep=0.2em]
      If \& I \& were \& at \& DreamWorks \& Animation \\
      SCONJ \& PRON \& AUX \& ADP \& PROPN \& PROPN \\
   \end{deptext}
   \depedge{3}{2}{nsubj}
   \depedge[arc angle=85]{3}{1}{mark}
   \depedge{3}{4}{prep}
   \depedge{4}{6}{pobj}
   \depedge{6}{5}{compound}
\end{dependency}
\end{center}
\caption{Antecedent in a counterfactual statement: Antecedent extends from `if' to the rightmost leaf node of sub-tree rooted at `were'} 
\end{figure}

\section{Implementation Details and Results}
\label{sec:result}

All the scores reported in the paper are on the validation set created by us by randomly splitting the dataset in a ratio of 75\%-25\%. We used the same training and validation sets across models to make sure the results were comparable. Please note we only report results of the key experiments - more details can be found at the GitHub repository. The results are reported in Tables \ref{table:res_1} and \ref{table:res_2}.

The highly imbalanced nature of the dataset posed a persistent challenge. We explored four methods to tackle it - simple oversampling of the minority class, SMOTE \cite{chawla2002smote}, undersampling the majority class, and using a weighted cross-entropy loss. The sampling was done such that the final number of positive and negative samples in the training set were equal. The weights in the cross-entropy loss were set to be the inverse of the class proportions i.e. mistakes on the minority class were penalised much more than those on the majority class. Out of all of these, we found that using the weighted cross-entropy loss provided the best results. All of the methods reported use this loss to tackle data imbalance.

For classical methods, we used the scikit-learn toolbox\footnote{\href{https://scikit-learn.org/stable/index.html}{https://scikit-learn.org/stable/index.html}} and spaCy to generate linguistic features. A linear SVM with the regularisation parameter set as 1.0 yielded the best results. The linear SVM has been observed to perform the best in sentence classification tasks \cite{joachims1998text}.

The implementation for the CNN based models in \cref{sssec:cnn} was kept similar to the original implementation by \cite{kim-2014-convolutional}. The convolutional kernel of sizes \{3,4,5\} was used with ReLU activation after each convolution and a max-pool layer after the final convolution. A fully-connected layer with the output being the number of classes (2 in our case) completed the model. A dropout \cite{srivastava2014dropout} layer was added before the final layer. An AdamW \cite{loshchilov2017decoupled} optimiser was used with an initial learning rate of $1 \times 10^{-3}$. As for the transformer models with a CNN head, the word-embedding matrix simply replaced the GloVe embeddings that were used in a stand-alone CNN model, with the rest of the model remaining the same.

The transformer-based models described in \cref{sssec:trans_s1} were implemented with the help of the Transformers library in PyTorch by HuggingFace\footnote{\href{https://github.com/huggingface/transformers}{https://github.com/huggingface/transformers}}. We used an AdamW optimiser for all of our models with an initial learning rate of $1 \times 10^{-5}$ over all the parameters and an epsilon of $1 \times 10^{-8}$. We trained the models for a total of 20 epochs and saved the model after each epoch if it provided the best F1 score on the validation set so far. We found that performance peaked around 16 epochs. We decayed the learning rate linearly to 0 over the training period. The choice of these involved quite a bit of fine-tuning as we found that results were sensitive to these hyper-parameters. The algorithm would not converge outside a specific range, and even within this range, the F1 could vary by 4-5 points upon small changes. 

As we mentioned earlier, the `large' versions of the transformers outperformed their `base' versions by 1-2\% in the case of BERT and 4-5\% in the case of the others. As shown, adding the CNN layer did actually improve performance in the relatively less-pretrained BERT, but slightly worsened performance on the sophisticated RoBERTa (large). Similarly, our experiments with linguistic features (LF) improved performance on base-sized transformers, but could not beat the performance of a fine-tuned large RoBERTa. The results for the discourse parsing-based (DP) method are also shown here - as mentioned earlier, they perform marginally (0.3\%-0.5\% F1) better than the models of \cref{sssec:trans_s1} b) and c).

\begin{table}
\begin{center}
\begin{tabular}{|l|c|c|c||l|c|c|c|}
\hline \bf Model & \bf Precision & \bf Recall  & \bf F1 & \bf Model  & \bf Precision & \bf Recall  & \bf F1 \\ \hline
Baseline (given) & 73.46 & 7.86 & 14.20 & SVM+3-gram & 64.64 & 55.89 & 59.95 \\
SVM+Tf-Idf & 77.45 & 34.49 & 47.73 & Above+POS & 73.60 & 58.37 & 65.10 \\
CNN (word2vec) & 64.41 & 75.78 & 69.63 & CNN (GloVe) & 69.18 & 75.50 & 72.20 \\ 
RoBERTa (large) & 88.98 & 88.98 & \bf 88.98 & XLNet (large) & \bf 92.96 & 83.75 & 88.12 \\
BERT (large) & 89.25 & 82.40 & 85.67 & BERT (b, CNN) &  89.05 & 85.12 & 87.04 \\
RoBERTa (l, CNN) & 89.82 & 82.64 & 86.08 & RoBERTa (b, LF) & 89.02 & 85.40 & 87.20 \\
RoBERTa (l, LF) & 86.01 & \bf 91.46 & 88.65 & SVM on `wish' & \bf 96.55 & 84.84 & \bf 90.32 \\
DP (Transformer) & 90.08 & 85.12 & 87.54 & DP (Word2Vec) & 69.83 & 71.36 & 70.59 \\
Vote (first)* & 85.96 & \b 92.83 & 89.27 & Vote (second)* & 90.04 & \bf 93.24 & \bf 91.18 \\

\hline
\end{tabular}
\end{center}
\caption{\label{table:res_1} Major results for subtask 1: classification of counterfactual statements. $*$ indicates those submitted for the competition. All results are on the same train/val/test split.}
\end{table}

For subtask 2, the transformer-based model in \cref{sssec:tree_ant} was implemented with the help of the Simple Transformers library\footnote{\href{https://github.com/ThilinaRajapakse/simpletransformers}{https://github.com/ThilinaRajapakse/simpletransformers}} in PyTorch. We tried BERT and RoBERTa as the models and used an AdamW optimiser with an initial learning rate of $4 \times 10^{-5}$ over all the parameters and an epsilon of $1 \times 10^{-8}$, and were trained for 5 epochs. We used spaCy for generating and visualising the dependency tree in \cref{sssec:tree_ant}.

\begin{table}
\begin{center}
\begin{tabular}{|l|c|c|c|c|}
\hline \bf Model & \bf Precision & \bf Recall  & \bf F1 & \bf Exact match \\ \hline
RoBERTa (large) & 45.80 & 48.10 & 44.90 & \bf 0.0349 \\
RoBERTa + DepParse* & \bf 48.30 & \bf 51.80 & \bf 47.10 & 0.0318 \\ 
\hline
\end{tabular}
\end{center}
\caption{\label{table:res_2} Major test results for subtask 2: identification of antecedent and consequent. $*$ indicates those submitted for the competition. All results are on the same train/val/test split.}
\end{table}

\section{Qualitative Analysis of Results}

Here we present a brief analysis of the classification results for Subtask 1. We try to identify some of the major reasons why our models perform the way they do, mostly by analysing the training data and the false predictions made by our best models.

Identifying whether a sentence is a counterfactual or not can be a difficult task even for humans - even we were often unable to agree upon whether a particular example in the training data was a counterfactual or not. However, one pattern is quite evident - many counterfactuals are conditionals e.g. of the form ``if... then...". Hence, the model learns to give a high weight to the embeddings corresponding to these words. Unfortunately, this means that a sentence having a similar structure is often falsely classified as a counterfactual. For example, the following statements were falsely classified as counterfactuals:
\begin{itemize}
	\item `\emph{Using simple math, you'd think that if you had worked 33 years and chose to work one more year, then you'd boost your benefits by about 1/33, or 3\%.}'
	\item  `\emph{Even if the Prime Minister's deal had been passed on Tuesday, there is a huge raft of legislation the Government would still need to pass.}'
\end{itemize} 

But the major issue in all the models was a low recall, i.e. a high number of false negatives. A low recall value was the major reason why we kept the voting threshold low (less than 50\%) for a statement to be classified as a counterfactual in our ensemble models. We believe that this was due to the models not being able to capture certain types of counterfactuals that do not follow a specific sentence structure. For example, consider the sentence: 

\begin{itemize}
\item `\emph{``Can you imagine if I said the things she said?" Mr. Trump told the crowd.}'
\end{itemize} 

This is a counterfactual, but our model was not able to focus on the word \textit{imagine} that makes it one. This shows that we either need a lot more data for different types of counterfactuals or a more objective set of rules that a model can be forced to learn. Sometimes a counterfactual is present as a part of a quote in a much more complex sentence such as
\begin{itemize}
\item `\emph{``So even if he can spend infinite money, it doesn't follow that he can use that money to condemn property," Somin said, adding that the administration would then be limited to building the predicted wall on land the federal government already own.}'
\item `\emph{It's become fashionable to tell a disability story in a hopeful arc, where the heroine may have moments of discouragement or fear, but comes out into full life at the end - into mainstream schools, love and romance, full participation in the social world, and these stories have become so pervasive that if they were to spread to aliens they'd find them familiar.}'
\end{itemize}
It is possible that the complexity and length of the sentence and the presence of many clauses that were clearly not counterfactual overshadowed the one clause that was a counterfactual and brought down the final score.

For subtask 2, the major challenge was that the antecedent and consequent which contributed to counterfactuality were often only parts of the longer sentence and hence it was difficult to conclude where the contributing bits ended. This problem was easier to solve in some cases where there was a pattern to the sentences as in the case of sentences containing `if'. There we saw that the dependency tree could be relied upon as there was a clear structure to the start of the antecedent, which depended on the main noun/verb of the clause.

\section{Conclusion and Future Work}

The identification and the analysis of counterfactual statements are essential for any computational natural language task, be it knowledge extraction or question answering. We attempt to identify such statements from a diverse corpus and produce competitive results using several SOTA language models. We modify some of the models and also use linguistic features to make our models task-specific, and with an ensemble of these models, we were able to rank 4th on the leaderboard in the first subtask. There is still scope for further improvements, such as hardcoding more specific rules in the model related to the various kinds of counterfactuals.

We also attempt to analyse the counterfactual statements by identifying the antecedent and the consequent. We model the problem as a sequence prediction problem and try out multiple transformer-based models as well as probabilistic ones. We ranked 9th on the leaderboard for this subtask. There are some other methods that one can experiment with for this task, such as separating the sentence into clauses and focusing on individual clauses, span-based models, analysing patterns involving other parts-of-speech, and graph-based networks that can leverage the power of the dependency tree correlation that we identified. We leave these methods for future work. 

\bibliographystyle{coling}
\bibliography{semeval2020}

\end{document}